\documentclass{INTERSPEECH2023}

\interspeechcameraready 
\usepackage{multirow}
\usepackage{booktabs}
\usepackage{graphicx}
\newcommand{\todo}[1]{}
\usepackage{amsmath}
\usepackage{subcaption}
\usepackage{url}
\usepackage{hyperref}
\usepackage{textcomp}
\usepackage[table]{xcolor}
\usepackage{bbding}
\renewcommand{\todo}[1]{{\color{red} [TODO: {#1}]}}

\definecolor{Gray}{gray}{0.9}

\title{Node-weighted Graph Convolutional Network for Depression Detection in Transcribed Clinical Interviews}

\name{Sergio Burdisso$^{\star,1}$, Esaú Villatoro-Tello$^{\star,1}$, Srikanth Madikeri$^1$, Petr Motlicek$^{1,2}$\thanks{This work was supported by Idiap internal funds. $^{\star}$\textit{Corresponding authors}.}}
\address{
  $^1$Idiap Research Institute, Switzerland\\
  $^2$ Faculty of Information Technology, Brno University of Technology, Czech Republic}
\email{\{sergio.burdisso, esau.villatoro, srikanth.madikeri, petr.motlicek\}@idiap.ch}

\begin{document}

\maketitle
 
\begin{abstract}

We propose a simple approach for weighting self-connecting edges in a Graph Convolutional Network (GCN) and show its impact on depression detection from transcribed clinical interviews. To this end, we use a GCN for modeling non-consecutive and long-distance semantics to classify the transcriptions into depressed or control subjects. The proposed method aims to mitigate the limiting assumptions of locality and the equal importance of self-connections vs. edges to neighboring nodes in GCNs, while preserving attractive features such as low computational cost, data agnostic, and interpretability capabilities. We perform an exhaustive evaluation in two benchmark datasets. Results show that our approach consistently outperforms the vanilla GCN model as well as previously reported results, achieving an F1=0.84 on both datasets. Finally, a qualitative analysis illustrates the interpretability capabilities of the proposed approach and its alignment with previous findings in psychology.

\end{abstract}
\noindent\textbf{Index Terms}: depression detection, graph neural networks, node weighted graphs, limited training data, interpretability.

\section{Introduction}
\label{sect:Introduction}



According to the World Health Organization (WHO), an estimated 970 million people in the world are living with a type of mental disorder, being depressive and anxiety disorders the most prevalent \cite{world2022depression}. Traditionally, the diagnosis and assessment for depression are done using semi-structured interviews and a Patient Health Questionnaire (PHQ) \cite{kroenke2009phq} as main tools, and it is generally based on the judgment of general practitioners.
However, practitioners may fail to recognize as many as half of all patients with depression \cite{mitchell2009}. Therefore, there is an acknowledged necessity for digital solutions for \textit{(i)} assisting practitioners in reducing misdiagnosis, and \textit{(ii)} addressing the burden of mental illness diagnosis and treatment \cite{wykes2019towards,vaidyam2019chatbots, Koulouri2022}.

Previous research has shown that language is a powerful indicator of our personality, social, or emotional status, and mental health \cite{pennebaker2003psychological, tackman2019depression}. As a result, many work exists at the intersection of artificial intelligence (AI), speech and natural language processing, psycholinguistics, and clinical psychology, showing that screening interviews, projective techniques, and essays writing provide valuable insights into the cognitive and behavioral functioning of subjects~\cite{malandrakis2015therapy, villatoroEtAl, villatoro2021applying, RAMIREZDELAROSA2023126}. 
Existing work on depression detection, via the use of textual transcriptions from psychotherapy sessions, varies from sentiment-based approaches \cite{10.3389/fpsyt.2021.811392}, going through methods designed to identify relevant vocabulary \cite{villatoroEtAl, burdisso2019text}, to various neural network architectures to best model the interviews, including bidirectional LSTM \cite{Li2022BidirectionalLA}, hierarchical attention-based networks \cite{Niu2021, Xezonaki2020AffectiveCO}, and deep neural graph structures \cite{hong2022using}. Other studies have experimented with multi-target hierarchical regression models to predict individual depression symptoms, aiming to improve performance by simultaneously predicting both binary diagnostic and depression severity regression scores \cite{Milintsevich2023TowardsAT}.
Finally, some works have explored the utility of enriching the models with additional (domain-specific) data \cite{Xezonaki2020AffectiveCO, zhang2020multimodal}, e.g., incorporating external linguistic knowledge to enforce higher values for attention weights corresponding to salient affective words.
Contrary to previous work, our proposed approach has the following salient features: does not require any external resource (data agnostic), does not depend on large pre-trained language models to learn embeddings (low computational cost), and has interpretability capabilities by design, a must in AI-supported diagnosis.

In particular, we propose to use a Graph Convolutional Network (GCN) to classify the transcribed sessions between a therapist and a subject seeking medical attention.
Overall, the main contributions of this paper are: (1) a novel weighting approach for self-connection nodes to address the limiting assumptions of locality and the equal importance of self-connections vs. edges to neighboring nodes in GCNs; (2) to the best of our knowledge, we evaluate for the first time an inductive implementation of GCNs in the task of depression detection from transcribed interviews, outperforming previously published results on two benchmark datasets; and (3) we demonstrate the interpretability potential of the proposed model, a key characteristic in AI-supported diagnosis, showing that what the model learned aligns with findings in psychology research.\footnote{Our code is available at \url{https://github.com/idiap/Node_weighted_GCN_for_depression_detection}} 

\section{Graph neural network architecture}
\label{sect:GCN_architecture}

\begin{figure}[t]
    \centering
    \includegraphics[width=\linewidth]{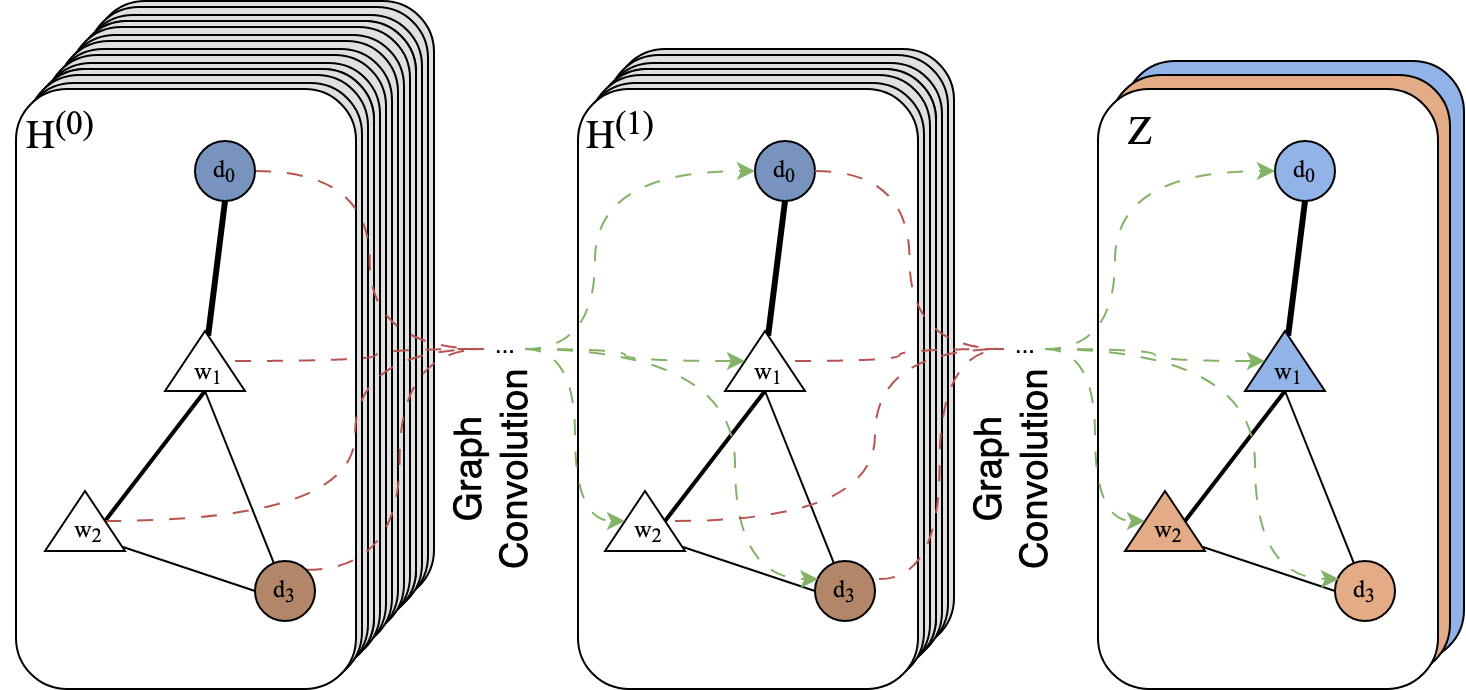}
    \caption{A two-layer GCN with nodes represented at three levels: initial (e.g. one-hot), $H^{(0)}$, intermediate/hidden, $H^{(1)}$, and output, $Z$, representations with the probability of each output label. Circles $\rightarrow$ document nodes \& triangles $\rightarrow$ word nodes.}
    \label{fig:GCN_architecture}
\end{figure}

A Graph Convolutional Network (GCN) is a multilayer neural network that operates directly on a graph and induces embedding vectors of nodes based on the properties of their neighbors \cite{yao2019graph, kipfsemi} (Figure \ref{fig:GCN_architecture}). Formally, considering a graph $G=(V, E, A)$, where $V (|V|=n)$ represents the set of nodes, $E$ is the set of edges, and $A\in \mathcal{R}^{n\times n}$ an adjacency matrix representing the edge values between nodes. The propagation rule for learning the new $k$-dimensional node feature matrix $H^{(l)}\in \mathcal{R}^{n\times k}$ is computed as:
\begin{equation}
    H^{(l+1)} = f(H^{(l)},A) = \sigma(\tilde{A}H^{(l)}W^{(l)})
\end{equation}
where $\tilde{A}=D^{-\frac{1}{2}}AD^{-\frac{1}{2}}$ represents the normalized symmetric adjacency matrix, $D_{ii} = \sum_{j}A_{ij}$ is the degree matrix of adjacency matrix $A$,  $W^{(l)}$ depicts the weight to be learned in the $l_{th}$ layer, and $\sigma$ is an activation function, e.g., ReLU: $\sigma(x)=\max(0,x)$.
In order to use GCNs for text classification \cite{yao2019graph}, we generate a large and heterogeneous text graph that contains word nodes ($V_{words}$) and \textit{training} document nodes ($V_{tr\_docs}$) so that global word co-occurrences can be explicitly modeled. 
Accordingly, the entire set of nodes is composed as $V=\{V_{tr\_docs}, V_{words}\}$, i.e. the number of training documents (corpus size) plus the number of unique words (vocabulary size) of the corpus. Particularly, in this work, we use a two-layer GCN defined as:
\begin{equation}
    H^{(1)}= \sigma(\tilde{A}H^{(0)}W^{(0)})    
\end{equation}
\begin{equation}
    Z = \text{softmax}(\tilde{A}H^{(1)}W^{(1)})
\end{equation}
where $W^{(0)}$ is the learned word embeddings lookup table, and $W^{(1)}$ represents the learned weight matrix in the second layer. Loss is computed by means of the cross-entropy function between $Z_{i}$ and $Y_{i}, \forall i \in V_{tr\_docs}$.  Intuitively the first layer learns the intermediate representation of the nodes (words and documents) while the second one learns the output representation, as illustrated in Figure \ref{fig:GCN_architecture}.
Note that in the output representation, label information from the documents has been propagated to the word nodes as output probabilities, allowing the model to learn the relation between words and output labels (e.g. depression or control labels), a key aspect favoring the interpretability of the model (see Section \ref{sec:interpretability}).

In order to make a fair comparison of the GCN's performance against other classification approaches, 
in this work we use the inductive version of GCNs as described in \cite{wang2022induct} instead of the original transductive one \cite{yao2019graph}.
Thus, the initial node feature matrix $H^{(0)}$ is generated such that word node vectors are represented as one-hot vectors, i.e., $H^{(0)}_i = \{0,1\}^{m}, \forall i \in V_{words}$, where $m$ is the vocabulary size of the training documents. And, for the representation of document node vectors $H^{(0)}_{i},\forall i \in V_{tr\_docs}$ the \emph{term-frequency-inverse document frequency} (TF-IDF) values of the corresponding word in that specific document is used, i.e., $H^{(0)}_{ij}= \text{TF-IDF}(i,j), \forall i,j$ where $i$ and $j$ are a document and a word, respectively.

For the definition of the edge types in $A$, we consider \textit{(i)} word-to-word, \textit{(ii)} word-to-document, similar to \cite{yao2019graph,wang2022induct}. Our key contribution here is the addition of a new edge type for \textit{(iii)} self-connections, acting as a trade-off parameter in the definition of $\tilde{A}$. 
Formally, this is expressed as follows:
\begin{equation}
\label{eq:edge_types}
A_{ij} =
    \begin{cases}
      \text{PMI}(i,j) & \text{if $i,j$ are words \& \text{PMI}$(i,j)>0$}\\
      \text{PR}(i,j) & \text{if $i,j$ are words \& $i=j$}\\
      \text{TF-IDF}_{i,j} & \text{if $i$ is document \& $j$ is word} \\
      0 & \text{otherwise}
    \end{cases}    
\end{equation}
where PMI is the Point-wise Mutual Information and PR stands for the \emph{PageRank} algorithm \cite{brin1998anatomy}, which given a graph computes the importance of each node in relation to the role it plays on the overall structure of the graph.
Intuitively, high PMI values will strongly link word nodes with high semantic correlation, high TF-IDF values will strongly link word nodes to specific document nodes, and high PageRank values will strongly link a node to itself proportionally to its global structural relevance;
this last modification aims to mitigate the assumption of locality and equal importance of self-loops, a known limitation in the vanilla GCN \cite{kipfsemi}. We will refer to this  modification as $\omega$-GCN.

Finally, it is worth mentioning that GCNs allow to easily optimize the model efficiency by means of applying simple feature selection techniques to reduce the vocabulary size (i.e. number of word nodes), prior to the graph construction, which has a direct impact on both the number of trainable parameters and model's interpretability (see section \ref{sec:models} and \ref{sec:interpretability}).
\section{Experimental setup}
\subsection{Datasets}
\label{sec:dataset}

For the experiments, we use the Distress Analysis Interview Corpus - wizard of Oz (DAIC-WOZ) dataset \cite{gratch-etal-2014-distress} and the Extended Distress Analysis Interview Corpus (E-DAIC) \cite{devault2014simsensei}. Both datasets contain semi-structured clinical interviews in North American English, performed by an animated virtual interviewer,\footnote{For DAIC-WOZ the virtual interviewer is human-controlled, while for the E-DAIC the virtual interviewer is fully automatic. A portion of the DAIC-WOZ transcriptions were generated using the ELAN tool from the Max Planck Institute for Psycholinguistics \cite{brugman-russel-2004-annotating}, while the E-DAIC transcripts were obtained using Google Cloud's ASR service.} designed to support the diagnosis of different psychological distress conditions. Datasets are multimodal corpora, composed by audio and video recordings, transcribed text from the interviews, and the Patient Health Questionnaire (PHQ-8 \cite{kroenke2009phq}) scores. During our experiments, we only used the speech transcripts from the subjects's responses.

Table \ref{tab:dataset} shows the composition of the datasets. Observe that the vocabulary size of the DAIC-WOZ is smaller than the E-DAIC vocabulary; suggesting a lesser variation of terminology in the provided answers, also reflected in a lower lexical richness (LR), an indicator of the E-DAIC complexity.

\begin{table}[t!]
    \scriptsize
    \centering
    \caption{Composition of the DAIC-WOZ and E-DAIC datasets for depressed (\textit{D}) and control (\textit{C}) participants. Column `Category' depicts the number of participants for each class, `Vocabulary' represents the vocabulary size for each partition, `LR' indicates the average lexical richness per instance, and `Duration' indicates the length (hrs:mins:secs) values.} 
    \label{tab:dataset}    
    \begin{tabular}{ccccccc}
        \toprule
         \rowcolor{Gray} \multicolumn{2}{c}{\textbf{Dataset}}& \multicolumn{2}{c}{\textbf{Category}} & \textbf{Vocabulary} & \textbf{LR} &\textbf{Duration}\\                  
        \midrule
         \multirow{4}{*}{\rotatebox{90}{DAIC-WOZ}} & \multirow{2}{*}{\textit{train}}& [D] 30 (\textit{28\%}) & &$m=5858$ & \multirow{2}{*}{0.48} &26h53m\\
         & & [C] 77 (\textit{72\%}) & &($\bar{x}$=621.11) &  &($\bar{x}$=15m04s)\\
        \cmidrule(r){2-7}
         & \multirow{2}{*}{\textit{dev}}& [D] 12 (\textit{34\%}) && $m=3268$ & \multirow{2}{*}{0.47} &10h01m\\
         & & [C] 23 (\textit{66\%}) &&  ($\bar{x}$=664.22) &  &($\bar{x}$=17m09s)\\
         \midrule
         \multirow{6}{*}{\rotatebox{90}{E-DAIC}}& \multirow{2}{*}{\textit{train}} & [D] 37 (\textit{23\%}) && $m=7991$ & \multirow{2}{*}{0.55} &43h29m\\
         & &[C] 126 (\textit{77\%}) && ($\bar{x}$=576.20) &  &($\bar{x}$=16m04s)\\
         \cmidrule(r){2-7}
         & \multirow{2}{*}{\textit{dev}}& [D] 12 (\textit{21\%}) && $m=4201$ & \multirow{2}{*}{0.58} &14h47m\\
         & & [C] 44 (\textit{79\%}) && ($\bar{x}$= 488.05) &  &($\bar{x}$=15m50s)\\
         \cmidrule(r){2-7}
         & \multirow{2}{*}{\textit{test}}& [D] 17 (\textit{30\%}) && $m=4183$ & \multirow{2}{*}{0.63} &15h14m\\
         & & [C] 39 (\textit{70\%}) && ($\bar{x}$=447.87) &  &($\bar{x}$=16m19s)\\
         \bottomrule
    \end{tabular}    
\end{table}


\subsection{Implementation details}
\label{sec:models}

As baseline models, we used different BERT-based models as well as simple models. More precisely, we used six pre-trained transformer-based models (\emph{bert-base-cased}, \emph{bert-base-uncased}, \emph{bert-large-cased}, \emph{bert-large-uncased}, \emph{roberta-base}, \emph{roberta-large}) to which a final linear layer was added to classify the input using, as usual, the \emph{[CLS]} classification special token.
In addition, to make the baselines as standard and simple as possible we made use of the \emph{Transformers} Python package~\cite{wolf2020transformers} \emph{AutoModelForSequenceClassification} class so that the size and number of linear layers are automatically selected according to each model. For each model, we also evaluated two versions, one enabling fine tuning of the base model and another not fine tuning the base model as part of the training process.
Regarding simple and classic models, we used a Support Vector Machine (SVM) with linear kernel and Logistic Regression (LR) model, both using TF-IDF-weighted words as features.

For GCN models, the size of nodes' intermediate representation was set to 64, i.e. we set $k=64$ for the $k$-dimensional feature matrix $H^{(1)}\in \mathcal{R}^{n\times k}$.
We performed a preliminary evaluation varying $k \in \{32, 64, 128, 256, 300\}$ from which $64$ showed to consistently be the best performing one.
In addition, since GCN models allow us to control the vocabulary size (i.e. number of word nodes), we trained different GCNs using different vocabulary sizes, as with SVM and LR models.
Namely, we applied the following feature selection techniques to build the vocabulary: (a) automatic selection based on term weights learned using LR; (b) top-$k$ best selection based on \emph{ANOVA F-value} between words and labels with $k \in \{100, 250, 500, 1000, 1500\}$; and (c) full vocabulary.
Trying different sizes allowed to control the complexity of the final model; GCNs with smaller vocabularies have smaller graphs, making them simpler and easier to interpret.

Finally, all neural-based models were implemented using PyTorch while non-neural ones using Scikit-learn. Additionally, for a fair comparison, all the models were optimized on each dataset using \emph{Optuna} \cite{akiba2019optuna} with 100 trials for hyperparameter search maximizing the macro averaged F1 score. For all neural-based models AdamW~\cite{loshchilov2018decoupled} optimizer ($\beta_1{=}0.9, \beta_2{=}0.999, \epsilon{=}1\mathrm{e}{-8}$) was used with \emph{learning rate} and number of epochs $n$ searched in $\gamma \in [1\mathrm{e}{-7}, 1\mathrm{e}{-3}]$ and $n \in [1, 10]$, respectively.
On the other hand, for non-neural baselines, search was performed varying the regularization parameter $C \in [1\mathrm{e}{-3}, 10]$, the class weight (balanced, none) and the penalty norm (L2, L1, L2 + L1, or none).
As a result, a total of 40 optimized models were  obtained.\footnote{14 simple baselines (SVM and LR with 7 vocabulary sizes), 12 BERT-based baselines (6 models with/without fine tuning), and 14 GCN models (vanilla GCN and $\omega$-GCN with 7 vocabulary sizes).}

\begin{table}[t!]
    \centering
    \scriptsize
    \caption{Results for \textit{dev} and \textit{test} partitions for DAIC-WOZ and E-DAIC datasets respectively. Performance is reported in terms of the F score ($F1$) for both control (\textit{C}) and depression (\textit{D}) classes, and the Macro-F for the overall problem (\textit{Avg.}).}
    \label{tab:results}
    \begin{tabular}{l@{~~~~~}c@{~~~~~}cccc}
    \toprule
    \multirow{2}{*}{\textbf{Method}} &  \multirow{2}{*}{\textbf{\#Params}} & \textbf{Vocab}& \multicolumn{3}{c}{\textbf{F1 score}}\\
    \cmidrule(lr){4-6}
     & & \textbf{size}& \textit{Avg.} & \textit{D} & \textit{C}\\
    \midrule
    \rowcolor{Gray} \multicolumn{6}{l}{\textbf{\quad \textit{DAIC-WOZ -- (dev)}}} \\
    \midrule 
    SVM  & 1952 & 1952 & 0.65 & 0.50 & 0.80\\
    LR   & 250 & 250 & 0.60  & 0.45 & 0.75\\    
    BERT & 335M & 30522 & 0.68 & 0.58 & 0.78\\
    BERT+FT & 335M & 30522 & 0.59 & 0.53 & 0.65\\    
    HCAG \cite{Niu2021} & - & - & 0.77 & - & -\\
    HAN-L \cite{Xezonaki2020AffectiveCO} & - & - & 0.69 & - & -\\
    Symptom-based \cite{Milintsevich2023TowardsAT} & - & - & 0.75 & - & -\\
    IDLV \cite{villatoroEtAl} & - & 100 & 0.64 & 0.52 & 0.77\\
    
    \cmidrule(lr){1-6}
    \textit{vanilla}-GCN & 375K & 5858 &0.75 & 0.67 & 0.83\\
    $\omega$-GCN & 375K & 5858 & 0.76 & 0.67 & 0.86\\
    \textit{vanilla}-GCN & 125K & 1952 &0.68 & 0.67 & 0.70\\
    $\omega$-GCN & 125K & 1952 & \textbf{0.79} & \textbf{0.76} & \textbf{0.83}\\
    \textbf{$\omega$-GCN$^\dag$} & 16K & 250 & \textbf{0.84} & \textbf{0.80} & \textbf{0.89}\\
    \midrule
    \rowcolor{Gray} \multicolumn{6}{l}{\textbf{\quad \textit{E-DAIC -- (dev)}}} \\
    \midrule 
    SVM & 7991 & 7991 & 0.69 & 0.47 & 0.91\\
    LR  & 7991 & 7991 & 0.71 & 0.53 & 0.90\\
    BERT & 108M  & 28996 & 0.61 & 0.46 & 0.75\\
    BERT+FT & 108M &28996 & 0.70 &0.54 & 0.86\\    
    IDLV \cite{villatoroEtAl} & - & 1000 & 0.64 & 0.38 & 0.90\\
    PV-DM \cite{zhang2020multimodal}$^*$ & - & - &0.90 &- &- \\
    \cmidrule(lr){1-6}
    \textit{vanilla}-GCN & 511K & 7991 & 0.71	& 0.50 &0.92\\
    \textbf{$\omega$-GCN} & 511K & 7991 & \textbf{0.80} & \textbf{0.67} & \textbf{0.94}\\
    \textit{vanilla}-GCN & 172K &2689 & 0.58 & 0.33 & 0.82\\
    $\omega$-GCN & 172K & 2698 & 0.70 & 0.54 & 0.86\\
    $\omega$-GCN$^\dag$ & 16K & 250 & 0.64	& 0.43 & 0.85\\
    \midrule
    \rowcolor{Gray} \multicolumn{6}{l}{\textbf{\quad \textit{E-DAIC -- (test)}}} \\
    \midrule 
    SVM & 250 & 250 & 0.69 & 0.60 & 0.78\\
    LR & 250 & 250 & 0.72 & 0.63 & 0.81\\
    BERT & 108M & 28996 & 0.49 & 0.29 & 0.696\\
    BERT+FT & 108M & 28996 & 0.75 & 0.65 & 0.85\\    
    VADER  \cite{10.3389/fpsyt.2021.811392} & - & - & - & 0.72 & 0.85 \\
    \cmidrule(lr){1-6}
    \textit{vanilla}-GCN & 511K & 7991 & 0.73	& 0.63 & 0.83\\
    $\omega$-GCN & 511K & 7991 & 0.72 & 0.63 & 0.81\\
    \textit{vanilla}-GCN & 172K & 2689 & 0.68 & 0.62 & 0.75\\
    $\omega$-GCN & 172K &2698 & 0.73	& 0.63 & 0.83\\
    \textbf{$\omega$-GCN$^\dag$} & 16K & 250 & \textbf{0.84} & \textbf{0.76} & \textbf{0.92}\\
    \bottomrule
    \end{tabular}
\end{table}

\section{Results}
\begin{figure*}[t!]
    \begin{subfigure}{0.5\textwidth}
        \centering
        \includegraphics[width=85mm]{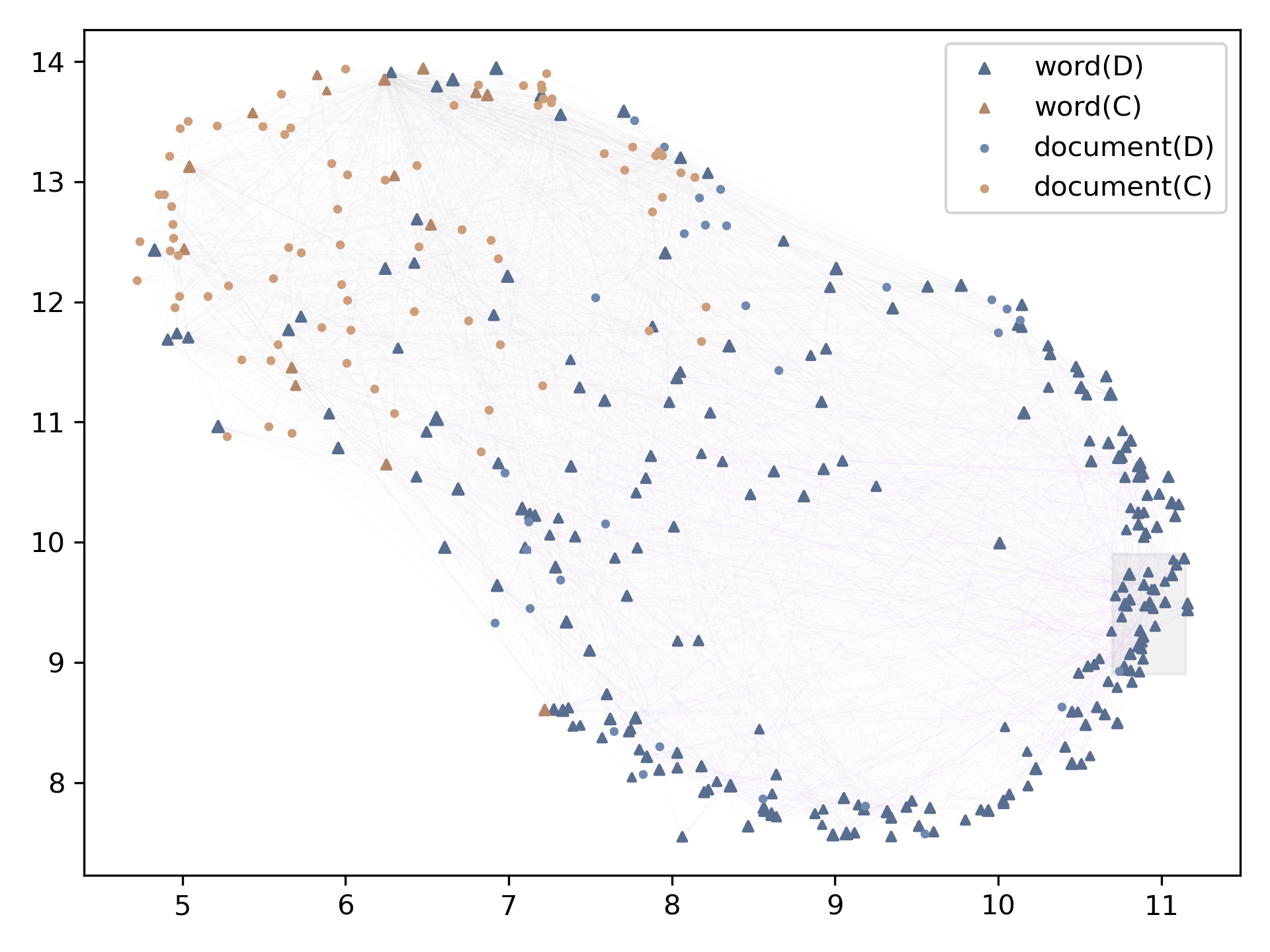}
        \caption{Overall graph with learned node embeddings}
        \label{fig:graph_250_16}
    \end{subfigure}
    \begin{subfigure}{0.5\textwidth}
        \centering
        \includegraphics[width=85mm]{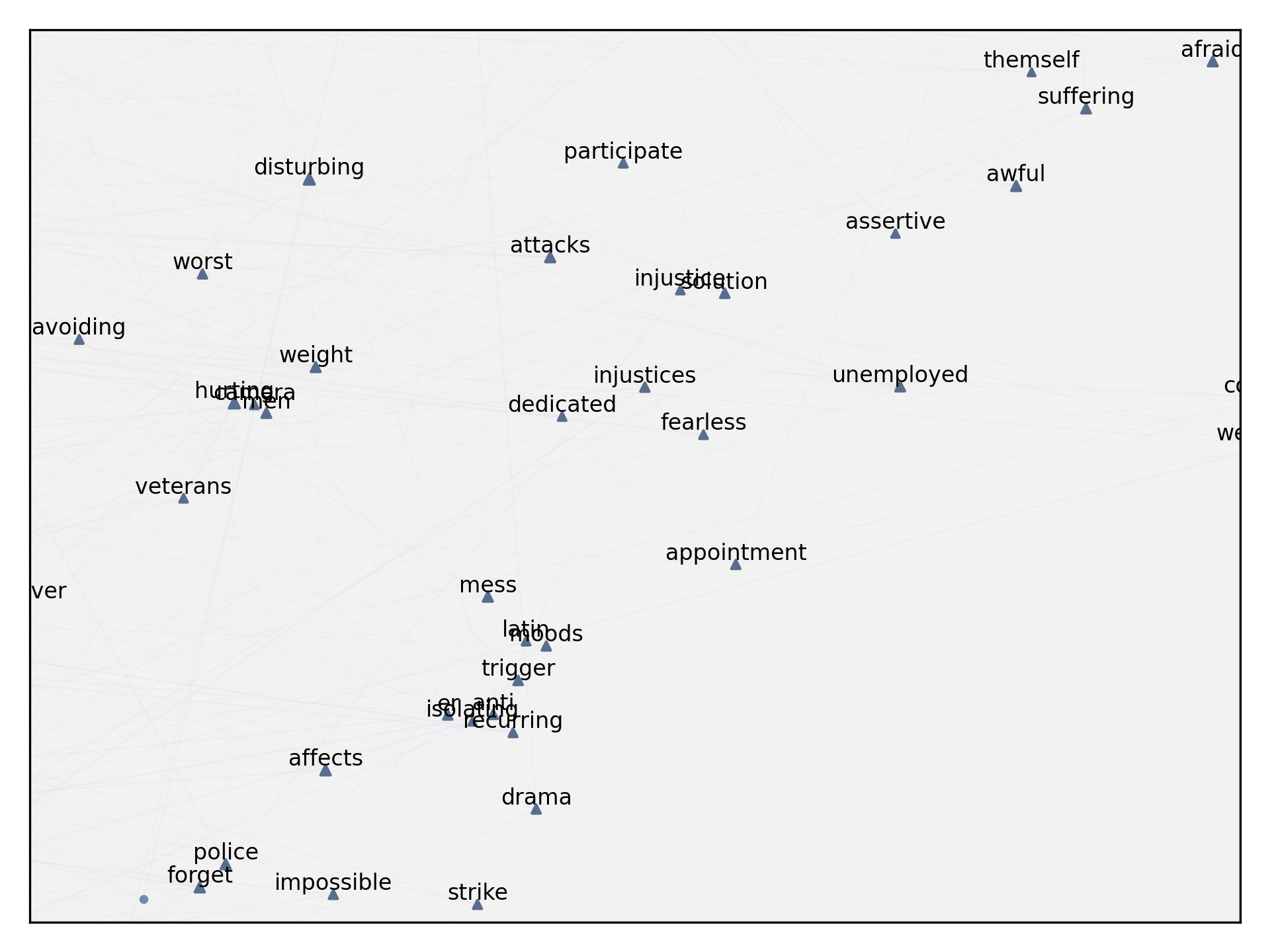}
        \caption{Zoomed-in region showing clusters of words (embeddings)}
        \label{fig:word_embeddings_zoom}
    \end{subfigure}
\caption{Node embbedings learned for DAIC-WOZ. As in Figure \ref{fig:GCN_architecture}, circles denote documents, triangles words, and colors denote class ([D] - depression, [C] - control). The gray rectangle in  (a) indicates the zoomed region (b). Graph edges are also included.}
\label{fig:graph_250}
\end{figure*}

Table \ref{tab:results} summarizes our results for the experiments on the \textit{dev} partition DAIC-WOZ and on the \textit{dev} and \textit{test} partitions of E-DAIC.\footnote{DAIC-WOZ \textit{test} partition is not publicly available.} For each partition, we divide the table into non-GCN models (i.e., classic and BERT-based baselines and previous research) and GCN models (vanilla GCN and our proposed $\omega$-GCN). In addition to the results, we also report the total number of trainable parameters (`\#Params') and the vocabulary size (`Vocab size'). Dashes indicate the corresponding metric is not reported in the original paper, while results marked with $*$ are not directly comparable as the model uses external domain-specific resources. Finally, for each dataset, we only report the best-performing models among all 40 optimized models (see Section \ref{sec:models}).

Overall, we see that the $\omega$-GCN approach consistently outperforms its vanilla version.
In addition, the model can outperform baselines and previously reported works when the correct number of features is selected.
For instance, on DAIC-WOZ, $\omega$-GCN obtains a macro $F1=0.84$ with only top-250 words. 
On the E-DAIC dataset, the $\omega$-GCN obtains the best performance among the considered methods, with a macro-$F1$ of $0.80$ and $0.84$ for the \textit{dev} and \textit{test} partitions respectively.
However, unlike the DAIC-WOZ \textit{dev} results, reducing the vocabulary size leads to unstable performance between \textit{dev} and \textit{test} sets suggesting models are sensitive to the (reduced) vocabulary discrepancy between the training and evaluation sets, a similar phenomenon as the one reported in \cite{villatoroEtAl}, where authors argue is due to the complexity of the dataset. We leave exploring methods to mitigate this phenomenon as future work by moving from a purely word-based vocabulary to, for instance, an embedding-powered or sub-word one (e.g. as BERT with WordPiece).

Finally, GCNs have order-of-magnitude fewer parameters than BERT models and are not constrained to a maximum sequence length (e.g. 512 tokens for BERT-based models).

\subsection{Exploring the model's interpretability}
\label{sec:interpretability}

One of the main advantages of the proposed GCN-based approach is that does not sacrifices performance for the sake of transparency.
Figure \ref{fig:graph_250} shows the \emph{UMAP}~\cite{McInnes2018} 2-dimensional projection of the 64-dimensional word and document embbedings learned by the best performing $\omega$-GCN model on DAIC-WOZ. 
More precisely, these embeddings correspond to the intermediate representation $H^{(1)}$, with the 250 word nodes painted with the learned class in the output representation $Z$.
The figure illustrates how the model can make use of the graph structure to learn, in the same latent space, document and word embeddings whose distance is influenced by their mutual relation and the output values.
These embeddings allow to identify clusters of strongly related words with high co-occurrence and linked to similar documents in the dataset, i.e., dataset-specific ``topics'' that experts could potentially use for qualitative analysis.
For instance, in DAIC-WOZ, interviews were conducted with war veterans and Figure \ref{fig:word_embeddings_zoom} depicts a few examples of these word clusters ---e.g. (1) about ``veterans'' and words like ``worst'', ``disturbing'', ``avoiding'' and ``hurting''; (2) about ``police'', ``strike'', ``drama'', ``moods'', ``trigger'', ``affects'', ``moods''; (3) about ``attacks'', ``injustices'', ``solution''; and (4) about ``unemployed'', ``suffering'', ``awful'', ``afraid''.

\begin{figure}[t!]
    \centering
    \includegraphics[width=\linewidth]{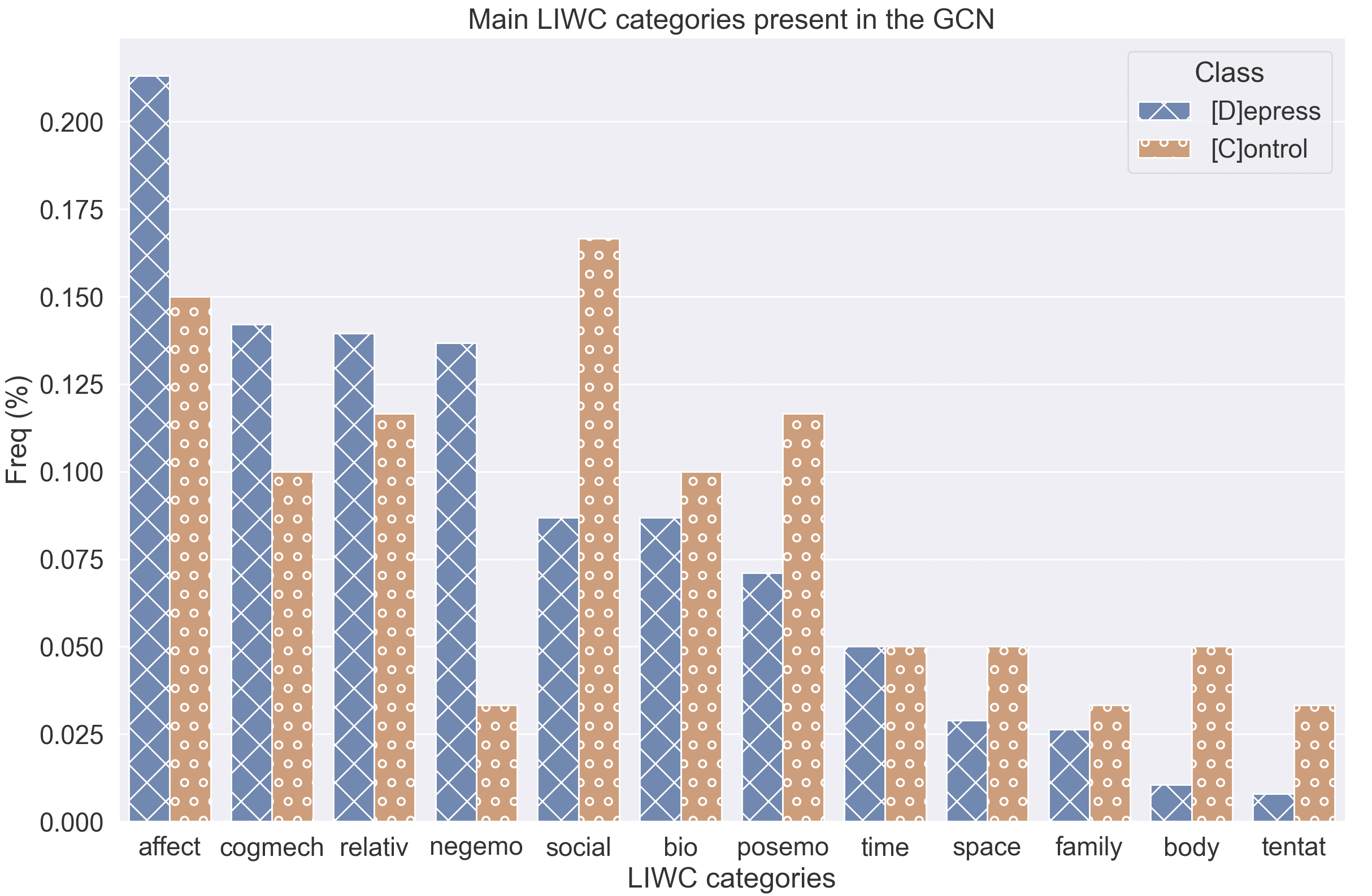}
    \caption{Psychological dimensions present in the model.}
    \label{fig:LIWC_analysis}
\end{figure}


Finally, we performed an analysis of how much of the acquired knowledge by the model fulfills known classical psychological theories/properties.
For this, we used the Linguistic Inquiry and Word Count (LIWC) \cite{pennebaker2001linguistic} lexical resource, composed of more than 4000 words, categorized into 64 psychological dimensions. Figure \ref{fig:LIWC_analysis} shows the result of this analysis. X-axis depicts the psychological dimensions of the words learned by the model, while the Y-axis represents the normalized frequency of the respective dimensions. As shown, the model learned that depressed subjects employ higher frequency dimensions related to affective or emotional processes (affect), cognitive processes (cogmech), relativity (relativ), and negative emotions (negemo). On the contrary, control subjects use more frequently the social processes (social), biological processes (bio), positive emotions (posemo), family and body dimensions. Overall, these findings are aligned with previously reported psychological work \cite{Mor2002}.


\section{Conclusions}
\label{sec:conclusions}
This paper proposes the use of Graph Convolutional Networks to detect depression from transcribed clinical interviews. The proposed approach has some attractive features, including a simple yet novel weighting approach for self-connection edges, a significantly low computational cost in terms of trainable parameters, and interpretability capabilities that help to understand the model's rationale.
Evaluation results on two depression-related datasets indicate that the proposed approach is able to consistently outperform its vanilla version. Our best configurations require orders of magnitude fewer trainable parameters than transformer-based models and yet, with the right vocabulary size, are able to obtain better F1 scores than baselines and previously reported results.
Finally, an exploration of the interpretability capabilities of the model showed that what it learned from raw data was, in fact, aligned with previously reported work from the psychological theory. 
As future work, we plan to use different nodes, from simple sub-word nodes to node hierarchies with different types. For instance, the addition of acoustic nodes, as a third type of node, would allow information transfer among acoustic, words and document embeddings.
\newpage
\bibliographystyle{IEEEtran}
\bibliography{mybib}

\end{document}